\documentclass[runningheads]{llncs}
\usepackage[T1]{fontenc}
\usepackage{graphicx}
\usepackage{booktabs}
\usepackage[misc]{ifsym}
\newcommand{\corr}{(\Letter)}
\usepackage[dvipsnames]{xcolor}
\usepackage{amsmath}
\usepackage{stackengine}
\usepackage{natbib}
\usepackage{hyperref}
\setcitestyle{authoryear,round, notesep={; }} 
\usepackage{mwe}
\usepackage{amsfonts}
\usepackage{upgreek}
\usepackage{multirow}
\usepackage{enumitem}
\usepackage{amsmath,amssymb}
\usepackage{tabularray}
\usepackage{svg}
\newcommand{\E}{\mathbb{E}}
\newcommand{\R}{\mathbb{R}}
\newcommand{\e}{\mathbf{e}}
\newcommand{\Th}{\Uptheta}
\newcommand{\agg}{\text{\rm \textsc{agg}}}
\newcommand{\upd}{\text{\rm \textsc{update}}}
\newcommand{\gnc}{\text{\rm \textsc{gnc}\space}}
\newcommand{\anc}{\text{\rm \textsc{anc}\space}}

\definecolor{dark2green}{rgb}{0.1, 0.65, 0.3}
\definecolor{dark2orange}{rgb}{0.9, 0.4, 0.}
\definecolor{dark2purple}{rgb}{0.4, 0.4, 0.8}
\newcommand{\first}[1]{\textbf{\textcolor{dark2green}{#1}}}
\newcommand{\second}[1]{\textbf{\textcolor{dark2orange}{#1}}}
\newcommand{\third}[1]{\textbf{\textcolor{dark2purple}{#1}}}

\begin{document}
\title{Next Level Message-Passing with Hierarchical Support Graphs}

\author{Carlos Vonessen$^*$ \and
Florian Grötschla$^*$ \corr \and
Roger Wattenhofer}

\authorrunning{Carlos Vonessen et al.}

\institute{ETH Zurich, Zurich, Switzerland\\ \email{\{cvonessen,fgroetschla,wattenhofer\}@ethz.ch}}

\maketitle

\def\thefootnote{*}\footnotetext{These authors contributed equally.}\def\thefootnote{\arabic{footnote}}

\begin{abstract}

Message-Passing Neural Networks (MPNNs) are extensively employed in graph learning tasks but suffer from limitations such as the restricted scope of information exchange, by being confined to neighboring nodes during each round of message passing. Various strategies have been proposed to address these limitations, including incorporating virtual nodes to facilitate global information exchange. In this study, we introduce the Hierarchical Support Graph (HSG), an extension of the virtual node concept created through recursive coarsening of the original graph. This approach provides a flexible framework for enhancing information flow in graphs, independent of the specific MPNN layers utilized. We present a theoretical analysis of HSGs, investigate their empirical performance, and demonstrate that HSGs can surpass other methods augmented with virtual nodes, achieving state-of-the-art results across multiple datasets.\footnote{Our implementation is available at \href{https://github.com/carlosinator/support-graphs}{https://github.com/carlosinator/support-graphs}}

\end{abstract}

\section{Introduction}
Graph Neural Networks (GNNs) have emerged as the leading method for learning on graph-structured data, through their ability to capture complex relationships between nodes. At the core of this field are Message-Passing Neural Networks (MPNNs), which iteratively exchange information between neighboring nodes through edges. Despite their success, MPNNs face significant limitations, primarily due to their restricted receptive field. 
This constraint hinders the ability of MPNNs to capture long-range dependencies effectively. To overcome the limited receptive field of MPNNs several methods have been proposed. A common strategy is to rewire the graph and enhance message-passing by creating additional edges. 
Another prominent approach is adding a virtual node that connects to all nodes in the graph, thus enabling global information exchange. This method has shown promise in extending the capabilities of MPNNs without significantly altering a graph's underlying structure.

In parallel, Graph Transformers (GTs), have gained traction by bypassing many of the inherent problems of MPNNs. GTs allow nodes to attend to all other nodes simultaneously, thereby eliminating the locality constraint of message passing.
As a result, GTs have demonstrated superior performance across various benchmarks. However, due to their asymptotic complexity, the scalability of GTs remains a challenge. GTs with full attention are computationally expensive and often impractical for large-scale graphs. Scalable variants of GTs, though more efficient, make other tradeoffs to stay performant.

We propose enhancing global information exchange in MPNNs and alleviating information bottlenecks by generalizing the concept of virtual nodes to general support structures we call Hierarchical Support Graphs (HSGs). 
In contrast to GTs, this approach only adds a small computational overhead and has the potential to be scaled to much larger graphs. The HSG is constructed through recursive coarsening of the original graph, providing a multi-level framework that enhances information exchange while maintaining computational efficiency. Our method integrates seamlessly with existing MPNN architectures, offering a scalable solution that bridges the gap between local and global information propagation.
This study presents a comprehensive theoretical analysis of HSGs, evaluates their empirical performance, and demonstrates their superiority over existing virtual node-augmented methods. Our results show that HSGs achieve state-of-the-art performance on several benchmark datasets, highlighting their potential as a robust and scalable enhancement for graph learning tasks. 

\section{Related Work}

\subsubsection{Limitations of MPNNs.}
Multiple works have investigated the limitations of standard MPNNs. \cite{alon2021bottleneck} and \cite{topping2022understanding} show that MPNNs struggle to transmit information through bottlenecks and along long paths. Further work has shown that simple MPNN configurations are only as expressive as the 1-Weisfeiler-Leman graph isomorphism test \citep{xu2019powerful}. \cite{oono2021graph} prove that on sufficiently dense graphs, simple GNNs with too many layers asymptotically lose their expressive power.

\subsubsection{Augmented Message-Passing.}
Several techniques have been proposed that augment the graph used for message-passing to improve information flow.
Virtual nodes (VNs), which connect to all nodes in a graph, are one of the most notable approaches~\citep{pham2017graph}. There have also been efforts to prove that VNs can improve the expressiveness of GNNs~\citep{cai2023connection,southern2024vnbron}. Most recently, \cite{rosenbluth2024distinguished} show that VNs can reach state-of-the-art performance on some datasets. 
Instead of only adding edges to newly introduced nodes, one can add or remove edges. This is usually referred to as graph rewiring, and various edge selection methods exist, ranging from curvature-based analysis~\citep{topping2022understanding,arnaiz2022diffwire} to dynamic rewiring~\citep{gutteridge2023drew} and the combination of rewiring with VNs~\citep{qian2024probabilistic}.
Hierarchical structures have also been used for domain-specific tasks~\citep{grotschla2022hierarchical}.
In contrast to previous work, HSGs offer general support structures that do not change node updates and only require minimal pre- and post-processing adaptation. Furthermore, we do not need to adapt the synchronous message-passing rounds, simplifying the architecture and implementation. 

\subsubsection{Graph Transformers.}
Transformers on graphs have recently gained traction, as they outperform traditional MPNNs on many datasets. They further alleviate many shortcomings of MPNNs by allowing all nodes to attend to each other simultaneously. \cite{rampasek2023graphgps} present GraphGPS, a scalable architecture that employs sparse graph attention and message passing to combine the advantages of either architecture. \cite{ma2023grit} show that adding inductive biases to graph transformers eliminates the need for message-passing modules. Several architectures, such as GraphGPS \citep{rampasek2023graphgps} or Exphormer~\citep{shirzad2023exphormer}, only require sparse attention that scales linearly in the number of nodes and edges. Hierarchical approaches have also been introduced for Graph Transformers~\citep{zhang2022ansgt}.

\subsubsection{Graph Coarsenings.}
Several works introduce coarsening and hierarchical clustering for learning on graphs. \cite{Chiang2019clustergcn} use graph clustering to identify well-connected subgraphs on large graphs. They show that it suffices to load these well-connected neighborhoods into memory rather than learning on the entire graph.
\cite{zhu2023hsgt,zhang2022ansgt} show that one can use hierarchical clustering to improve the scalability of GTs on large-scale graphs.
Similarly, \cite{huang2021scaling} create coarser representations of large-scale graphs and train a GNN only on the coarser representation, yielding message-passing with sublinear complexity.
\cite{ying2019hierarchical} recursively coarsen graphs using learned node embeddings to generate graph classification predictions. 
\cite{groetschla2024coregd} use graph coarsening to compute representations for graph drawing efficiently.
Further works also employ graph coarsening to improve learning on graphs \citep{sobolevsky2021hierarchical,fang2020hierarchical,bergmeister2024efficient}.\\
In contrast to previous works, we propose a generalizable and flexible approach by integrating the HSG into the original graph. We further use hierarchical clustering to improve long-range information exchange in MPNNs rather than to decrease the asymptotic complexity of GTs as done in previous works \citep{zhang2022ansgt,zhu2023hsgt,kuang2022coarformer}.

\section{Preliminaries}
\subsection{MPNNs}
\label{sec:mpnn-intro}
In this section, we introduce standard MPNNs and MPNNs augmented with a virtual node. We first state some elementary definitions.
\begin{definition}
    Given an undirected graph $G$ with node-set $V(G)$ and edge-set $E(G)$, we define the following
    \begin{enumerate}
        \item Let $\mathcal N_G(v):=\{u\in V(G)\;|\; \{u,v\}\in E(G)\}$ be the neighborhood of a vertex $v\in V(G)$.
        \item The vector $h_v^{(t)}$ denotes the hidden representation of a node after message-passing round $t$. $h_v^{(0)}$ is the initial node feature.
        \item We denote $\agg$ as an aggregation module that combines a multiset of vectors into one.
        \item We denote {\rm \textsc{update}} as a function that combines two vectors into one.
    \end{enumerate}
    When it is clear which graph is being referenced we shorten the notation to $V$ and $E$.
\end{definition}
\begin{definition}[Message-Passing Layer]
\label{def:MPNN}
Given a graph $G$ we define a message-passing (MP) module as
\begin{align}
    m_v^{(t+1)} &= \agg(\{ \! \{h_u^{(t)}\mid u\in \mathcal N_G(v)\} \! \}),\\
    h_v^{(t+1)} &= \upd(h_v^{(t)}, m_v^{(t+1)}),
\end{align}
\noindent where $\{ \! \{\cdot\} \! \}$ denotes a multi-set.
\end{definition}
An MPNN typically consists of an array of MP modules executed in series. Additional task-specific pre- and post-processing is often done with multi-layer perceptrons (MLP). A common MP module is the Graph Convolutional Network (GCN), introduced by \cite{kipf2017gcnconv}. One can extend the definition of MP modules to take edge features into account in the \upd{} step. A popular example of this is the GatedGCN architecture \citep{bresson2018gatedgcn}. 
In node property prediction tasks, one can directly use a node's hidden representation after an array of MP modules as input to an MLP to generate a prediction. In graph property prediction tasks, a common approach is to pool all nodes with a global pooling function to produce one unified graph representation.

\subsection{Graph Measures}
\label{sec:graph-meas}
We introduce several common graph measures that we will use to empirically assess the impact of HSG augmentation on graph topology. 

The notion of effective resistance on graphs is derived from electric circuit analysis, where one considers all edges in a graph to be resistors. In this work, we consider all edges to have a resistance of 1. Intuitively two nodes will have low effective resistance if they are connected either through a few short paths or a larger amount of longer paths. \\
The following theorem defines the graph resistance for two nodes. While it is unknown who first proved it, a proof can be found in \cite{kleinResistanceDistance1993}.
\begin{theorem}[Effective Resistance]
    Given a graph $G$, let $D \in \R^{|V|\times|V|}$ be the diagonal matrix containing the node degrees and $A\in \R^{|V|\times|V|}$ be the adjacency matrix. Then $L=D-A$ defines the graph Laplacian. Furthermore, $L^+$ denotes the Moore-Penrose pseudoinverse of $L$ and $\e_x$ the all zero vector with a 1 at position $x$. \\
    Given a graph $G$ one can compute the effective resistance $R_{ab}$ between a pair of nodes $a,b\in V$ as
    \begin{equation}
        R_{ab} = (\e_a - \e_b)L^+(\e_a - \e_b)\;.
    \end{equation}
\end{theorem}

\cite{kleinResistanceDistance1993} also prove that the effective resistance is a distance measure and can thus be used to measure how well two nodes are connected. It is thus intuitive to use the effective resistance to model how well two nodes can communicate in an MPNN. \\

The commute time in a random walk on a graph is defined as the expected number of steps necessary to reach $b$ from $a$ and return.
\begin{definition}[Hitting Time]
    In a random walk $(X_t)_{t\in\mathbb N_0}$ on a graph $G$ with $a,b\in V(G)$ and $X_0=a$, the hitting time is the time required to reach $b$ from $a$.
    \begin{equation}
        H_{ab} = \inf\{t\in T\mid X_t=b\}\;.
    \end{equation}
    From the definition of the hitting time, one can define the commute time as $C_{ab} = H_{ab} + H_{ba}$.
\end{definition}

\begin{theorem}[Resistance and Commute Time~\citep{chandra1989electrical}]
\label{thm:reff-ct}
    In a graph $G$ it holds that for any node pair $a,b\in V$
    \begin{equation}
        \E[C_{ab}] = 2 \cdot |E| \cdot R_{ab}
    \end{equation}
\end{theorem}
It follows that the commute time scales proportionally to the effective resistance and the number of edges. We consider this measure in addition to the effective resistance, as we study the effect of the additional edges introduced by the HSG. \\

The connectivity of a node pair $u,v\in V$ is defined as the number of node disjoint paths between $u$ and $v$. This is equivalent to the minimum number of nodes one must remove from the graph to disconnect $u$ and $v$. 

\begin{definition}[Node Connectivity]
\label{def:node-conn}
Let $\#\text{cc}$ denote the number of connected components of a graph, and $G[X]$ the subgraph induced by $X\subset V$. We define the graph node connectivity {\rm (\textsc{gnc})} as
\begin{equation}
\gnc := \min_{S\subset V} \{|S|\mid \#\text{cc}(G[V\textbackslash S]) > 1\}\;.
\end{equation}
We further define the connectivity between two nodes $u,v\in V$ (with $u\neq v$) as
\begin{align}
    c_{uv} = \min \{|S|\mid\text{\rm there exists no path between $u$ and $v$ in $G[V\textbackslash S]$}\}\;,
\end{align}
and the average node connectivity {\rm (\textsc{anc})} as the average over all $c_{uv}$.
\end{definition}

\anc is a useful measure to model the ability two nodes have to exchange information. In MPNNs, each node on a path only has a limited capacity to transmit information. We thus consider the number of node-disjoint paths between two nodes when assessing their ability to communicate.\\

\section{Hierarchical Support Graphs}
\begin{figure}[t]\scriptsize
    \centering
    \includegraphics[width=0.43\textwidth]{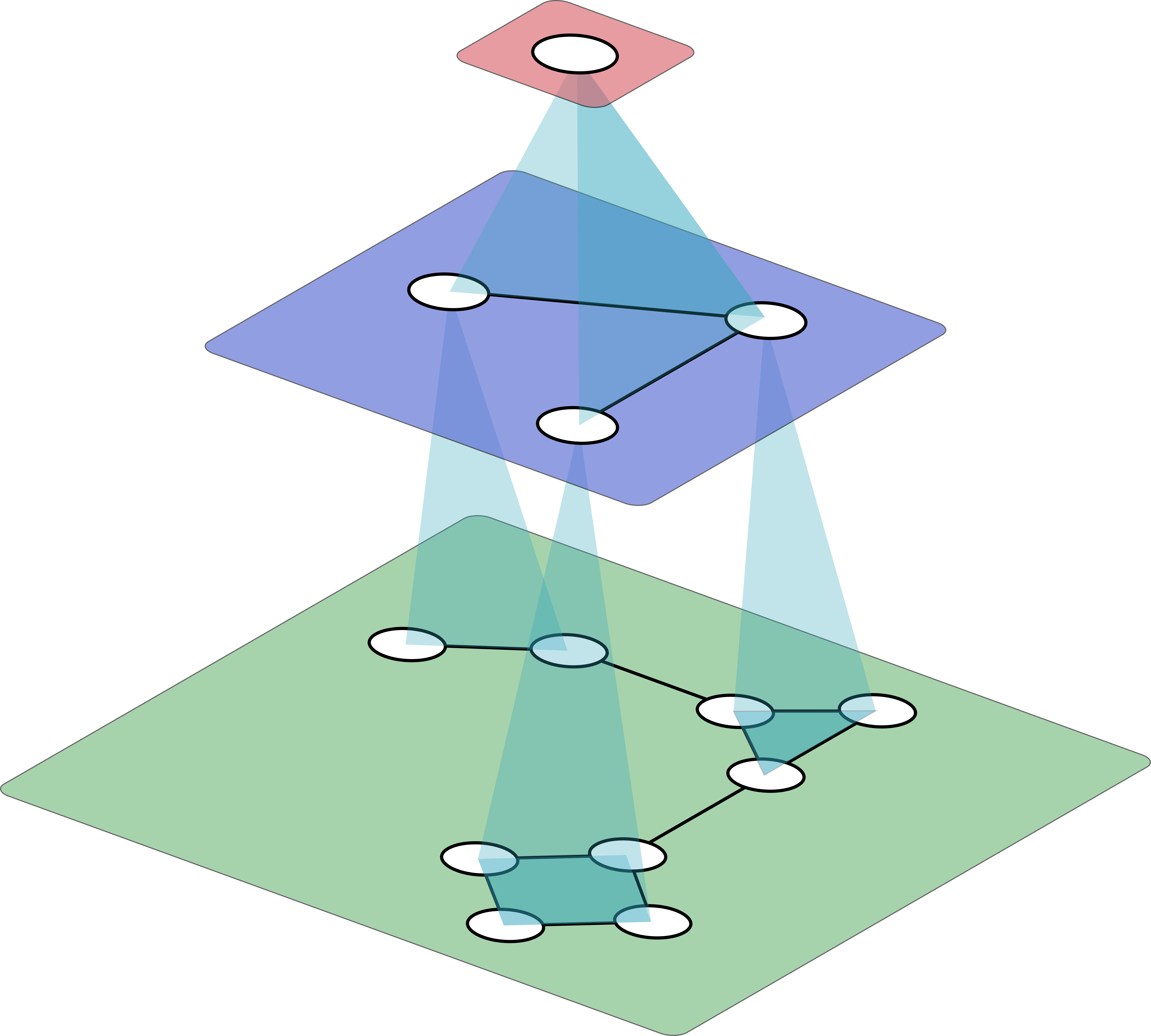}
    \hspace{1cm}
    \includegraphics[width=0.36\textwidth]{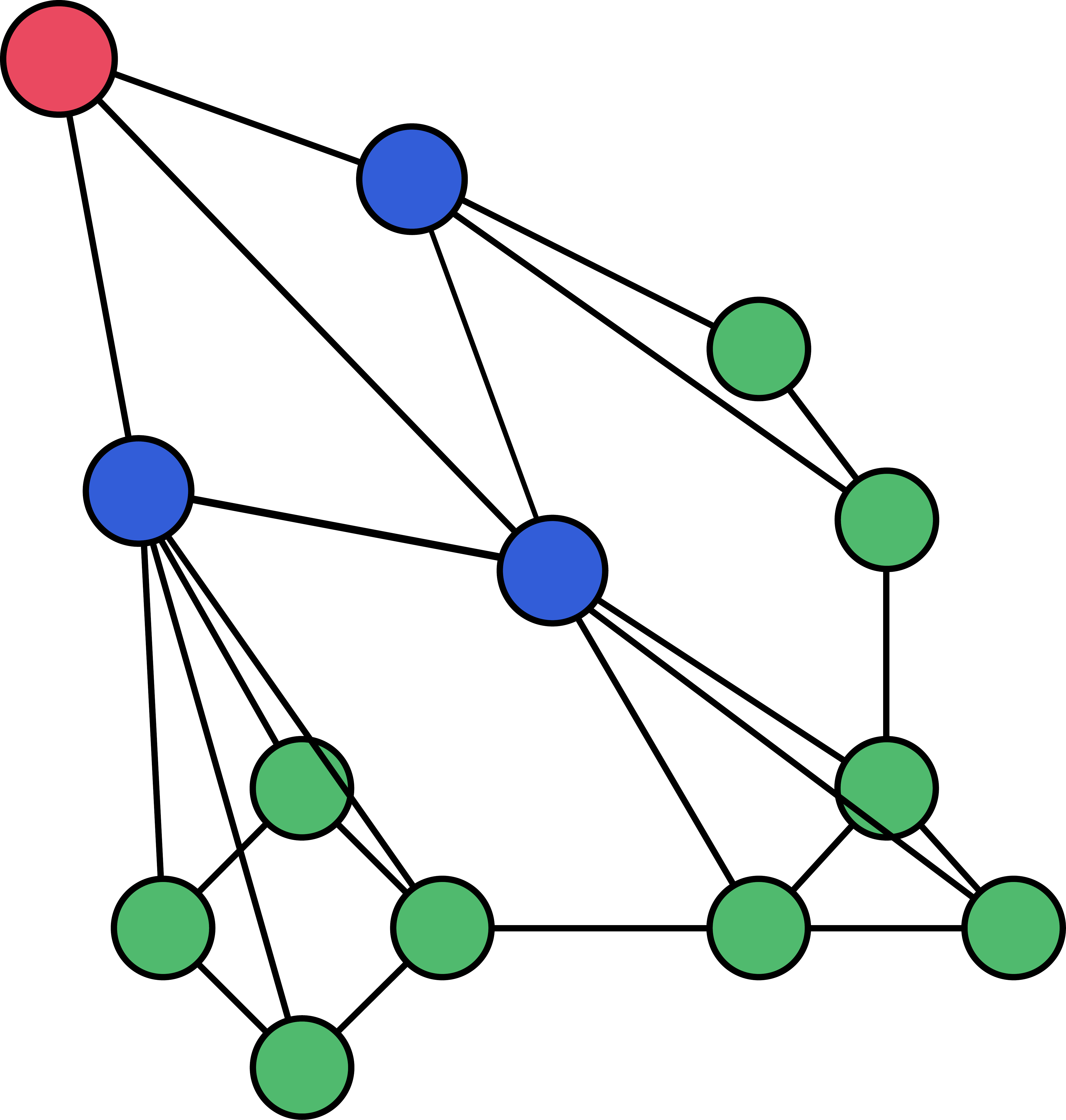}
    
    \caption{Visualization of the graph coarsening procedure. First, the input graph $G$ is recursively coarsened by clustering nodes in the same layer. Consequently, each node corresponds to a super-node representing the cluster. Finally, the super-nodes and edges are integrated as regular nodes into the graph.}
    \label{fig:how-to-coarsen}
\end{figure}

HSGs are created through recursive graph clustering. One takes the original graph and creates $k$ node clusters by, for example, minimizing the number of intercluster edges. These clusters are then contracted to $k$ super-nodes and multiple edges between clusters are contracted to single super-edges. We call this graph a support graph, as it mimics the structure of the original graph at a coarser level. \\
We repeat this process recursively, creating a hierarchy of increasingly coarser graphs in each layer. We denote edges within one support graph layer as horizontal edges. As a final step, we connect all support hierarchies and the original graph by letting each node connect to its direct super-node. We denote these edges as vertical edges. We visualize the procedure in Figure~\ref{fig:how-to-coarsen}.

\begin{definition}[Graph Coarsening]
    Let $G$ be an input graph. A coarsening algorithm with strength $r\in [0,1)$ computes a partitioning $\{K_1,\dots, K_{q}\}$ of the node set $V(G)$ with $q=\lfloor r\cdot n \rfloor$. From this, one can construct a new graph $H$ where each node $v_i\in H$ represents a node partition $K_i$ in the original graph $G$. Furthermore, one adds an edge $\{v_i, v_j\}\in E(H)$ if there exists at least one edge between the nodes in $K_i$ and $K_j$ in $G$. We refer to this process as $H = f_r(G)$.
\end{definition}

\begin{definition}[HSG Augmentation]
\label{def:basic-erg}
We apply a coarsening algorithm $f_r$ to the input $G$, yielding the first support graph layer: $H^{(1)} = f_r(G)$. We repeat this recursively $Z-1$ more times, with $H^{(i+1)} = f_r(H^{(i)})$. The new graph, which is a combination of the original graph $G$ and all support graph layers $(H^{(i)})_{i\in[Z]}$ is denoted as $G^H$.\\
For convenience, we denote the set of all nodes as $\hat{V} := \bigcup_{i\in[Z]} V(H^{(i)}) \cup V(G)$.
\begin{enumerate}
    \item We define $\varphi: \hat{V} \rightarrow \hat{V}$ as the map of any node $v\in \hat{V}$ to their direct super-node. The map is not defined for nodes in the highest layer $v\in H^{(Z)}$. \\
    \item We define the node-set and edge-set of the HSG-augmented graph $G^H$ as
    \begin{align}
        V(G^H) &:= \hat{V},\\
        E(G^H) &= \bigcup_{i\in[Z]} E(H^{(i)}) \cup \{\{v,\varphi(v)\}\mid\forall\; v\in \hat{V}\} \cup E(G)\;.
    \end{align}
\end{enumerate}
\end{definition}

This definition allows us to state some basic properties of HSG-augmented graphs.

\begin{theorem}[Appendix~\ref{sec:proof-basic-props}]
\label{thm:basic-props}
    Given a graph $G$ with $n$ nodes and $m$ edges. For any recursive coarsening method with a constant coarsening ratio $r\in(0,1]$, the total number of nodes in $G^H$ is tightly upper bounded by $\frac{n}{1-r}$. The diameter of $G^H$ is bounded by $2\frac{\log n}{-\log r}$. Similarly, the number of edges is at most $\mathcal{O}(\frac{m\log n}{-\log r})$.
\end{theorem}

Since the HSG is integrated into normal message-passing, each $H^{(i)}$ should preserve some properties of the original graph. For example, the node degree should not explode compared to the original nodes. In the following, we briefly examine this property and give bounds that coarsening methods should satisfy to fulfill this constraint.

\begin{definition}[Cumulative Coarsening]
    Let $G$ be a graph of $n$ vertices and $m$ edges. In the $i$-th coarsening step, we define $r(i)$ as the node coarsening strength. We further study $c(i)$, which is the edge reduction factor. We define both formally as
    \begin{align}
        r(i) &= \frac{|V(H^{(i-1)})|}{|V(H^{(i)})|}\;,\\
        c(i) &= \frac{|E(H^{(i-1)})|}{|E(H^{(i)})|}\;.
    \end{align}
    We denote $R(i):=\prod_{j\leq i} r(j)$ as the cumulative node reduction and $C(i):=\prod_{j\leq i} c(j)$ as the cumulative edge reduction compared to the original graph. Intuitively, $R(i)\cdot n$ is the number of nodes in the $i$-th hierarchical layer, and $C(i)\cdot m$ is the number of edges.
\end{definition}
For many coarsening algorithms one can control $r(i)$, however $c(i)$, the factor by which the number of edges is reduced, commonly cannot be chosen.
\begin{theorem}[Appendix~\ref{app:useful-coar}]
\label{thm:const-ratio}
    In a connected graph $G$ with $n$ vertices and $m$ edges, any coarsening method that preserves the average node degree of horizontal and vertical edges $\frac{m+n}{n}$ must satisfy $\forall i \leq Z$
    \begin{align}
        r(i)&=\Upomega\left(\frac{n}{m}\right),\\
        C(i)&=\Th(R(i))\,,
    \end{align}
    For $r(i)=\Th(\frac{n}{m})$ the condition relaxes to $C(i)=\mathcal{O}(R(i))$.
\end{theorem}

This bound on $C(i)$ restricts the types of coarsening that one can apply to approximately preserve the original average node degree. 
We analyze this restriction using a random coarsening on the Erdős–Rényi Random Graph model and prove that many Erdős–Rényi Graphs will, in expectation, preserve the original node degree up to constant factors.

\begin{theorem}
\label{thm:gnp-pvr}
    Given an Erdős–Rényi Graph $G(n,p)$, and a random coarsening with equal cluster sizes $(\pm 1)$ and coarsening ratio $r=\Th(\frac{n}{m+n})$. The node degree for horizontal edges remains unchanged up to constant factors compared to the previous layer's degree for
    \begin{equation}
        p(n) = \Th(n^{-\beta}),
    \end{equation}
    where $\beta\in\{\frac{1}{2}\}\cup[1,\infty)$. For $\beta\in[0, \frac{1}{2})$ the horizontal node degree is asymptotically smaller than the previous node degree. The horizontal node degree diverges for all other values of $\beta$.
\end{theorem}
From this, one can directly restrict the permissible region of $p$.
\begin{theorem}
\label{thm:gnp-whatgraphs}
    An Erdős–Rényi Graph $G(n,p)$ which, when coarsened with $r=\Th(\frac{n}{m+n})$, preserves the node degree of the previous layer up to constant factors can be restricted to
    \begin{equation}
    p(n) = \Th(n^{-\beta}),
    \end{equation}
    with $\beta\in[0, \frac{1}{2}]\cup[1,\infty)$.
\end{theorem}
For $p\geq \frac{(1+\epsilon)\ln n}{n}$ an Erdős–Rényi Graph will be connected with high probability \citep{erdosevolutionrandom}. In practice, Theorem~\ref{thm:gnp-whatgraphs} covers the edge density of many relevant graphs.

\subsection{Integrating HSGs into GNNs}
We generate HSGs using the METIS partitioning algorithm~\citep{karypis1998metis}. In each hier\-archical support layer, the algorithm attempts to minimize the number of inter-cluster edges while creating similar sized node clusters. After generating the HSG layers, we introduce the created nodes and edges as regular edges into the message-passing framework, thereby eliminating the need for any changes to MP modules. As a result, compared to several previous works that compute a separate update step for the virtual node \citep{rosenbluth2024distinguished,southern2024vnbron,cai2023connection}, our approach requires no custom layers or updates specific to the HSG. This also allows us to minimize the changes one has to make to an existing MPNN pipeline. In a pre-processing step we compute the coarsening and augment the input graph as a pre-transformation. Using imputation, one can compute features for the virtual nodes and edges a priori (see Section ~\ref{sec:ablation}). Finally, using simple node and edge indicators, one can completely recover the original topology and add specific embeddings to nodes and edges.

\subsubsection{Adapted Graph Pooling.} 
\label{sec:adapted-pooling}
We further make use of the highest HSG layer to adapt the kind of graph pooling we apply on graph-level tasks. We reduce the global pooling introduced in Section~\ref{sec:mpnn-intro} to only operate over the highest layer of HSG nodes $H^{(Z)}$, which then serves as input to the MLP prediction head. 

\section{Experiments}
We evaluate our approach on PascalVOC-SP, COCO-SP, Peptides-func, and Peptides-struct from the Long Range Graph Benchmark (LRGB) \citep{dwivedi2023longrange} and on ogbg-molpcba from the Open Graph Benchmark (OGB) \citep{hu2021ogb}. PascalVOC-SP and COCO-SP are vision datasets where each node represents a contracted area, or superpixel, from the original image. Peptides-func, -struct, and ogbg-molpcba are molecule datasets. Details on the model configurations are given in Appendix~\ref{app:model-configs}.

\subsection{How HSGs Change Graph Topology}
In this Section, we investigate the effect of HSG augmentation on real-world graphs. Due to its high computational cost, we restrict our analysis to 100 random graphs from Peptides-func. 
We compare several graph measures for unmodified graphs, graphs augmented with a virtual node, and graphs augmented with an HSG. We give average node properties such as pair-wise distances and \anc for the original node pairs and ignore virtual nodes or HSG nodes. 

\begin{table}\scriptsize
    \centering
    \caption{Graph statistics for a random subset of 100 graphs from Peptides-func. $\bullet$ represents the contraction into a single virtual node. For example, the configuration $(0.25, \bullet)$ denotes a coarsening method where one contracts to 25\% of the original graph and then contracts the new layer to one supernode.}
    \resizebox{\textwidth}{!}{%
    \begin{tabular}{cccccccccc}
        \toprule
        Augmentation & Coarsening & \stackanchor{Avg}{nodes}& \stackanchor{Avg}{edges}& Diameter & \stackanchor{Avg}{shortest path} & $\E[R_{ab}]$ & $\E[C_{ab}]$ & GNC & ANC \\
        \midrule
        original & - & $151.12$ & $153.93$ & $57.13$ & $20.90$ & $20.24$ & $7740.86$ & $1.00$ & $1.02$ \\
        virtual node & $(\bullet)$ & $152.12$ & $305.05$ & $2.00$ & $1.98$ & $0.90$ & $550.03$ & $2.00$ & $2.02$ \\
        \midrule
        \multirow{2}{*}{METIS} & $(0.25,\bullet)$ & $189.81$ & $380.22$ & $4.00$ & $3.81$ & $1.44$ & $1111.70$ & $2.00$ & $2.28$ \\
        & $(0.5,\bullet)$ & $207.99$ & $417.92$ & $4.00$ & $3.84$ & $1.40$ & $1181.68$ & $2.00$ & $2.37$ \\
        \midrule
        \multirow{2}{*}{Random} 
        & $(0.25,\bullet)$ & $189.31$ & $474.88$ & $4.00$ & $3.50$ & $0.96$ & $914.67$ & $2.00$ & $2.60$ \\
        & $(0.5,\bullet)$ & $217.59$ & $516.39$ & $4.00$ & $3.70$ & $1.01$ & $1050.26$ & $2.00$ & $2.60$  \\
        \bottomrule
    \end{tabular}
    }
    \label{tab:graph-stats}
\end{table}

Table~\ref{tab:graph-stats} shows both the effective resistance and the expected commute time between all original input nodes. We observe a drastic reduction for both measures in the VN and HSG-augmented graphs compared to the original graphs. 
We further observe that the additional hierarchical layer slightly increases both measures compared to the VN-graphs. This gap is slightly larger for commute times, as HSG-augmented graphs introduce more edges than a single VN. 
Furthermore, there is a significant increase in \anc compared to the virtual node, both for METIS and random coarsenings. 
It should be noted that \anc takes into account that nodes have a limited capacity for transmitting information, which might be of interest for certain tasks.
Finally, one can observe that random coarsenings exhibit lower commute time, effective resistance, and higher \anc compared to METIS coarsenings, likely because Peptides-func contains sparse graphs, and introducing a small number of random edges and nodes can aid connectivity.

\subsection{Evaluation on Benchmark Datasets}
Table~\ref{tab:results} shows the empirical results on all tested datasets. On LRGB, we give the values of SAN and GPS as reported in \cite{rosenbluth2024distinguished}. The values of SAN on ogbg-molpcba are reported from \cite{kreuzer2021rethinking}. We report the performance of DRew as given in \cite{gutteridge2023drew} and of GRIT as reported in \cite{ma2023grit}. The performance of GatedGCN-VN and GCN-VN is given as reported in \cite{rosenbluth2024distinguished}. Finally, we use the baselines of GCN and GatedGCN as reported in \cite{rosenbluth2024distinguished}. 
We note that DRew and SAN are evaluated without the improved feature normalization introduced by \cite{toenshoff2023wheredid}. We give the mean and standard deviation over 7 runs on ogbg-molpcba and over 4 runs over the LRGB datasets. Further details can be found in Appendix~\ref{app:model-configs}. 

\begin{table} \scriptsize
    \centering
    \caption{Test performance on four benchmarks from LRGB \citep{dwivedi2023longrange} and ogbg-molpcba \citep{hu2021ogb}. The runs mark the \first{first}, \second{second}, and \third{third} highest performing models.}
    \resizebox{\textwidth}{!}{%
    \begin{tabular}{cccccc}
        \toprule
        \multirow{2}{*}{Model} & PascalVOC-SP & COCO-SP & Peptides-func & Peptides-struct & ogbg-molpcba\\
        & F1 Score $\uparrow$ & F1 Score $\uparrow$ & AP $\uparrow$ & MAE $\downarrow$ & AP $\uparrow$\\
        \midrule
         GCN
         & $20.78\pm0.31$ & $13.38\pm0.07$ & $68.60\pm0.50$ & \third{$\mathbf{24.60\pm0.07}$} & $24.83\pm0.37$\\
         GatedGCN
         & $38.80\pm0.40$ & $29.22\pm0.18$ & $67.65\pm0.47$ & $24.77\pm0.09$ & \third{$\mathbf{30.66\pm0.13}$} \\
         \midrule
         SAN
         & $32.30 \pm 0.39$ & $25.92 \pm 1.58$ & $64.39 \pm 0.75$ & $25.45 \pm 0.12$ & $27.65 \pm 0.42$ \\
         GPS
         & \third{$\mathbf{44.40\pm0.65}$} & \first{$\mathbf{38.84 \pm 0.55}$} &  $65.35\pm0.41$ & $ 25.09 \pm 0.14$ & $29.07 \pm 0.28$\\
         GRIT
         & - & - & \third{$\mathbf{69.88 \pm 0.82}$} & \third{$\mathbf{24.60 \pm 0.12}$} & - \\
         \midrule
         DRew
         & $33.14\pm0.24$ & - & \second{$\mathbf{71.50\pm0.44}$} & $25.36\pm0.15$ & -\\
         S$^2$GCN
         & - & - & \first{$\mathbf{73.11\pm0.66}$} & \second{$\mathbf{24.47\pm0.32}$} & - \\
         \midrule
         GatedGCN-VN
         & \second{$\mathbf{44.7\pm1.37}$} & \third{$\mathbf{32.44\pm0.25}$} & $68.23\pm0.69$ & $24.75\pm0.18$ & \first{$\mathbf{31.41\pm0.19}$}\\
         GCN-VN
         & $29.5\pm 0.58$ & $20.72\pm0.43$ & $67.32\pm0.66$ & $25.05\pm0.22$ & - \\
         \midrule
         GatedGCN-HSG 
         & \first{$\mathbf{46.04\pm0.59}$} & \second{$\mathbf{35.35\pm0.32}$} & $68.66\pm0.38$ & \first{$\mathbf{24.21\pm0.07}$} & \second{$\mathbf{31.29\pm0.20}$}\\
         GCN-HSG 
         & $27.36\pm0.49$ & $18.89\pm0.42$ & $68.91\pm0.29$ & $24.79\pm0.07$ & $26.89\pm0.25$ \\
         \bottomrule
    \end{tabular}
    }
    \label{tab:results}
\end{table}

Table~\ref{tab:results} shows state-of-the-art performance on PascalVOC-SP and Peptides-struct. Furthermore, using GatedGCN, we outperform the virtual node implementations of \cite{rosenbluth2024distinguished}. We further highlight that our approach outperforms or matches GT approaches on PascalVOC-SP, Peptides-func, Peptides-struct, and ogbg-molpcba. On COCO-SP, GatedGCN-HSG shows significantly higher predictive performance than previous MPNN-based approaches and reduces the gap to GraphGPS.
The graph-level tasks Peptides-struct and ogbg-molpcba are trained using the adapted graph pooling introduced in Section~\ref{sec:adapted-pooling}. Furthermore, these two datasets are only augmented with the smallest possible HSG, a single virtual node. This configuration probably works best because the average graph size in ogbg-molpcba is 26~\citep{hu2021ogb}.
We further note that the two vision datasets PascalVOC-SP and COCO-SP outperform previous virtual node implementations using GatedGCN, but fail to match the corresponding implementation using the GCN module. As we show in Table~\ref{tab:imputation}, masking HSG-edges with dummy features significantly improves predictive performance. As GatedGCN uses edge gates to distinguish neighbors, the dummy features could help mark the super-nodes and -edges as such. In contrast, the lower performance of GCN compared to the virtual node implementations by \cite{rosenbluth2024distinguished} could be explained by GCN having no mechanism to gate edges in contrast to GatedGCN. \cite{rosenbluth2024distinguished} update the virtual node in a separate step, thus circumventing this issue. 

\subsection{Ablation studies} 
\label{sec:ablation}
In this Section, we investigate multiple hyperparameters of HSGs to evaluate their effect on predictive performance. We first observe how different coarsening configurations impact predictive capabilities. We further evaluate the impact of feature imputation on model performance. Finally, we investigate the benefits of HSG-adapted graph pooling.
\subsubsection{Coarsening Configurations.}
We investigate the effect of different coarsening configurations on a GCN-HSG model trained on Peptides-func. We show results in Table~\ref{tab:func-coarsen}. We observe only minor differences, as many configurations lie within one standard deviation of each other. Here, the virtual node is outperformed by METIS but performs marginally better than random coarsening. However, as the model configurations in Appendix~\ref{app:model-configs} indicate, in several cases, a virtual node integrated into message passing outperforms a larger METIS-coarsened HSG. We further observe a significant performance difference between METIS and random coarsening. Section~\ref{sec:graph-meas} shows that random coarsenings have the highest \anc and lowest effective resistance among HSG graphs, yet one can observe a clear performance gap in Table~\ref{tab:func-coarsen}. This may indicate that the graph structure, that is preserved through METIS coarsening is relevant to predictive performance.

\begin{table}[]
    \centering
    \caption{Average Precision of different coarsening configurations. The best-performing configuration is marked in \textbf{bold}.
    $(\bullet)$ corresponds to a virtual node integrated into normal message passing.}
    \begin{tabular}{ccc}
        \toprule
        Coarsen config & METIS & Random \\
        \midrule
        $(0.25, \bullet)$ & $68.71\pm0.60$ & $67.39\pm0.79$\\
        $(0.5, \bullet)$ & $\mathbf{68.91\pm0.29}$ & $67.92\pm0.46$\\
        $(\bullet)$ & \multicolumn{2}{c}{$68.19 \pm 0.62$}\\
        \bottomrule
    \end{tabular}
    \label{tab:func-coarsen}
\end{table}

\subsubsection{Feature Imputation.}
We investigate the effect of different feature imputation methods on the created nodes and edges of an HSG. We test different approaches on PascalVOC-SP with GatedGCN-HSG and on Peptides-func with GCN-HSG. We show results in Table~\ref{tab:imputation}. $\textbf{dummy-}$ denotes that nodes or edges only receive a dummy embedding denoting their hierarchical position in the graph.  $\textbf{impute-}$ denotes that the features of nodes or edges are imputed from their direct children. Based on what the features represent, we impute them using the mean in the vision dataset PascalVOC-SP and the mode in the molecule dataset Peptides-func. \\
GatedGCN-HSG trained on PascalVOC-SP profits from imputed nodes and dummy edges. The dummy edges especially may help the edge-gating mechanism in GatedGCN to distinguish real and hierarchical nodes. GCN-HSG trained on Peptides-func exhibits no performance difference for edge imputation as the GCN does not take edge features into account. This missing information could be compensated by clearly marking hierarchical nodes with dummy embeddings.

\begin{table}
    \centering
    \caption{Impact of super-node and -edge feature imputation for the PascalVOC-SP and Peptides-func datasets. The best-performing configuration is marked in \textbf{bold}.}
    \begin{tabular}{ccccc}
        \toprule
        \multirow{2}{*}{Dataset} & \multicolumn{2}{c}{PascalVOC-SP \hspace{0.25cm} \ } & \multicolumn{2}{c}{Peptides-func} \\
        & impute-node & dummy-node \hspace{0.25cm} \ & impute-node & dummy-node \\
        \midrule
        impute-edge \hspace{0.25cm} \ & $45.06\pm1.06$ & $44.19\pm0.37$\hspace{0.25cm} \ & $66.31\pm0.23$ & $68.88\pm0.86$ \\
        dummy-edge \hspace{0.25cm} \ & $\mathbf{46.04\pm0.59}$ & $43.81\pm0.95$\hspace{0.25cm} \ & $66.04\pm1.04$ & $\mathbf{68.91\pm0.29}$ \\
         \bottomrule
    \end{tabular}
    \label{tab:imputation}
\end{table}

\subsubsection{Adapting the Prediction Head.}
The HSG allows us to experiment with additional graph pooling methods on graph-level tasks. In Table~\ref{tab:custom-graph-head}, we compare the model performance when we apply global pooling and when we aggregate only the node features of the highest HSG level $H^{(Z)}$.\\
The results show that for ogbg-molpcba and Peptides-struct, only aggregating the highest layer of the HSG significantly improves performance. 
Compared to Peptides-func, both ogbg-molpcba and Peptides-struct only use a single virtual node as HSG augmentation, which means that global information must travel through the virtual node. Peptides-func is augmented with a two-layer HSG, thus information can also travel through lower layers.

\begin{table}\scriptsize
    \centering
    \caption{Performance on three graph-level datasets comparing predictive performance using global node pooling and top layer node pooling. The best configuration in each column is marked in \textbf{bold}.}
    \resizebox{\textwidth}{!}{%
    \begin{tabular}{cccc}
        \toprule
        Pred. Head & Peptides-func (AP $\uparrow$) & Peptides-struct (MAE $\downarrow$) & ogbg-molpcba (AP $\uparrow$) \\
        \midrule
        global pooling & $\mathbf{68.91\pm0.29}$ & $24.75\pm0.12$ & $30.14\pm0.24$ \\
        top layer pooling & $67.30\pm0.17$ & $\mathbf{24.21\pm0.07}$ & $\mathbf{31.29\pm0.20}$ \\
        \bottomrule
    \end{tabular}
    }
    \label{tab:custom-graph-head}
\end{table}

\section{Conclusion}
In this paper, we introduced Hierarchical Support Graphs (HSGs), an extension of the virtual node concept designed to enhance message-passing in graph neural networks by alleviating information bottlenecks. Through a combination of empirical and theoretical analyses, we demonstrate the impact of HSGs on graph topology and information exchange.
Our experiments on common benchmarking datasets further reveal that HSGs can achieve state-of-the-art performance, surpassing methods that rely solely on single virtual nodes, and achieving a new state-of-the-art on two datasets. 
In conclusion, HSGs offer a generalizable framework for improving message-passing neural networks that effectively extends the reach of information propagation without introducing substantial complexity. 

\begin{credits}
\subsubsection{\ackname} The authors acknowledge Leo Widmer for his assistance in resolving the proofs in Appendix~\ref{app:random-proof}.
\end{credits}

\bibliography{mybibliography}

\begin{thebibliography}{35}
\providecommand{\natexlab}[1]{#1}
\providecommand{\url}[1]{\texttt{#1}}
\expandafter\ifx\csname urlstyle\endcsname\relax
  \providecommand{\doi}[1]{doi: #1}\else
  \providecommand{\doi}{doi: \begingroup \urlstyle{rm}\Url}\fi

\bibitem[Alon and Yahav(2021)]{alon2021bottleneck}
U.~Alon and E.~Yahav.
\newblock On the bottleneck of graph neural networks and its practical implications, 2021.

\bibitem[Arnaiz-Rodr{\'\i}guez et~al.(2022)Arnaiz-Rodr{\'\i}guez, Begga, Escolano, and Oliver]{arnaiz2022diffwire}
A.~Arnaiz-Rodr{\'\i}guez, A.~Begga, F.~Escolano, and N.~Oliver.
\newblock Diffwire: Inductive graph rewiring via the lov$\backslash$'asz bound.
\newblock \emph{arXiv preprint arXiv:2206.07369}, 2022.

\bibitem[Bergmeister et~al.(2024)Bergmeister, Martinkus, Perraudin, and Wattenhofer]{bergmeister2024efficient}
A.~Bergmeister, K.~Martinkus, N.~Perraudin, and R.~Wattenhofer.
\newblock Efficient and scalable graph generation through iterative local expansion, 2024.

\bibitem[Bresson and Laurent(2018)]{bresson2018gatedgcn}
X.~Bresson and T.~Laurent.
\newblock Residual gated graph convnets, 2018.

\bibitem[Cai et~al.(2023)Cai, Hy, Yu, and Wang]{cai2023connection}
C.~Cai, T.~S. Hy, R.~Yu, and Y.~Wang.
\newblock On the connection between mpnn and graph transformer, 2023.

\bibitem[Chandra et~al.(1989)Chandra, Raghavan, Ruzzo, and Smolensky]{chandra1989electrical}
A.~K. Chandra, P.~Raghavan, W.~L. Ruzzo, and R.~Smolensky.
\newblock The electrical resistance of a graph captures its commute and cover times.
\newblock In \emph{Proceedings of the twenty-first annual ACM symposium on Theory of computing}, pages 574--586, 1989.

\bibitem[Chiang et~al.(2019)Chiang, Liu, Si, Li, Bengio, and Hsieh]{Chiang2019clustergcn}
W.-L. Chiang, X.~Liu, S.~Si, Y.~Li, S.~Bengio, and C.-J. Hsieh.
\newblock Cluster-gcn: An efficient algorithm for training deep and large graph convolutional networks.
\newblock In \emph{Proceedings of the 25th ACM SIGKDD International Conference on Knowledge Discovery and Data Mining}, KDD ’19. ACM, July 2019.
\newblock URL \url{http://dx.doi.org/10.1145/3292500.3330925}.

\bibitem[Dwivedi et~al.(2023)Dwivedi, Rampášek, Galkin, Parviz, Wolf, Luu, and Beaini]{dwivedi2023longrange}
V.~P. Dwivedi, L.~Rampášek, M.~Galkin, A.~Parviz, G.~Wolf, A.~T. Luu, and D.~Beaini.
\newblock Long range graph benchmark, 2023.

\bibitem[Erd{\H o}s and R{\'e}nyi(1960)]{erdosevolutionrandom}
P.~Erd{\H o}s and A.~R{\'e}nyi.
\newblock {{On the Evolution of Random Graphs}} by.
\newblock 1960.

\bibitem[Fang et~al.(2020)Fang, Sun, Gan, Pillai, Wang, and Liu]{fang2020hierarchical}
Y.~Fang, S.~Sun, Z.~Gan, R.~Pillai, S.~Wang, and J.~Liu.
\newblock Hierarchical graph network for multi-hop question answering, 2020.

\bibitem[Gr{\"o}tschla and Mathys(2022)]{grotschla2022hierarchical}
F.~Gr{\"o}tschla and J.~Mathys.
\newblock Hierarchical graph structures for congestion and eta prediction.
\newblock \emph{arXiv preprint arXiv:2211.11762}, 2022.

\bibitem[Grötschla et~al.(2024)Grötschla, Mathys, Veres, and Wattenhofer]{groetschla2024coregd}
F.~Grötschla, J.~Mathys, R.~Veres, and R.~Wattenhofer.
\newblock Core-gd: A hierarchical framework for scalable graph visualization with gnns, 2024.

\bibitem[Gutteridge et~al.(2023)Gutteridge, Dong, Bronstein, and Giovanni]{gutteridge2023drew}
B.~Gutteridge, X.~Dong, M.~Bronstein, and F.~D. Giovanni.
\newblock Drew: Dynamically rewired message passing with delay, 2023.

\bibitem[Hu et~al.(2021)Hu, Fey, Zitnik, Dong, Ren, Liu, Catasta, and Leskovec]{hu2021ogb}
W.~Hu, M.~Fey, M.~Zitnik, Y.~Dong, H.~Ren, B.~Liu, M.~Catasta, and J.~Leskovec.
\newblock Open graph benchmark: Datasets for machine learning on graphs, 2021.

\bibitem[Huang et~al.(2021)Huang, Zhang, Xi, Liu, and Zhou]{huang2021scaling}
Z.~Huang, S.~Zhang, C.~Xi, T.~Liu, and M.~Zhou.
\newblock Scaling up graph neural networks via graph coarsening, 2021.

\bibitem[Karypis and Kumar(1998)]{karypis1998metis}
G.~Karypis and V.~Kumar.
\newblock A fast and high quality multilevel scheme for partitioning irregular graphs.
\newblock \emph{SIAM Journal on Scientific Computing}, 20\penalty0 (1):\penalty0 359--392, 1998.
\newblock URL \url{https://doi.org/10.1137/S1064827595287997}.

\bibitem[Kipf and Welling(2017)]{kipf2017gcnconv}
T.~N. Kipf and M.~Welling.
\newblock Semi-supervised classification with graph convolutional networks, 2017.

\bibitem[Klein and Randi{\'c}(1993)]{kleinResistanceDistance1993}
D.~J. Klein and M.~Randi{\'c}.
\newblock Resistance distance.
\newblock \emph{Journal of Mathematical Chemistry}, 12\penalty0 (1):\penalty0 81--95, Dec. 1993.
\newblock ISSN 1572-8897.
\newblock \doi{10.1007/BF01164627}.

\bibitem[Kreuzer et~al.(2021)Kreuzer, Beaini, Hamilton, Létourneau, and Tossou]{kreuzer2021rethinking}
D.~Kreuzer, D.~Beaini, W.~L. Hamilton, V.~Létourneau, and P.~Tossou.
\newblock Rethinking graph transformers with spectral attention, 2021.

\bibitem[Kuang et~al.(2022)Kuang, WANG, Li, Wei, and Ding]{kuang2022coarformer}
W.~Kuang, Z.~WANG, Y.~Li, Z.~Wei, and B.~Ding.
\newblock Coarformer: Transformer for large graph via graph coarsening, 2022.
\newblock URL \url{https://openreview.net/forum?id=fkjO_FKVzw}.

\bibitem[Ma et~al.(2023)Ma, Lin, Lim, Romero-Soriano, Dokania, Coates, Torr, and Lim]{ma2023grit}
L.~Ma, C.~Lin, D.~Lim, A.~Romero-Soriano, P.~K. Dokania, M.~Coates, P.~Torr, and S.-N. Lim.
\newblock Graph inductive biases in transformers without message passing, 2023.

\bibitem[Oono and Suzuki(2021)]{oono2021graph}
K.~Oono and T.~Suzuki.
\newblock Graph neural networks exponentially lose expressive power for node classification, 2021.

\bibitem[Pham et~al.(2017)Pham, Tran, Dam, and Venkatesh]{pham2017graph}
T.~Pham, T.~Tran, H.~Dam, and S.~Venkatesh.
\newblock Graph classification via deep learning with virtual nodes, 2017.

\bibitem[Qian et~al.(2024)Qian, Manolache, Morris, and Niepert]{qian2024probabilistic}
C.~Qian, A.~Manolache, C.~Morris, and M.~Niepert.
\newblock Probabilistic graph rewiring via virtual nodes.
\newblock \emph{arXiv preprint arXiv:2405.17311}, 2024.

\bibitem[Rampášek et~al.(2023)Rampášek, Galkin, Dwivedi, Luu, Wolf, and Beaini]{rampasek2023graphgps}
L.~Rampášek, M.~Galkin, V.~P. Dwivedi, A.~T. Luu, G.~Wolf, and D.~Beaini.
\newblock Recipe for a general, powerful, scalable graph transformer, 2023.

\bibitem[Rosenbluth et~al.(2024)Rosenbluth, Tönshoff, Ritzert, Kisin, and Grohe]{rosenbluth2024distinguished}
E.~Rosenbluth, J.~Tönshoff, M.~Ritzert, B.~Kisin, and M.~Grohe.
\newblock Distinguished in uniform: Self attention vs. virtual nodes, 2024.

\bibitem[Shirzad et~al.(2023)Shirzad, Velingker, Venkatachalam, Sutherland, and Sinop]{shirzad2023exphormer}
H.~Shirzad, A.~Velingker, B.~Venkatachalam, D.~J. Sutherland, and A.~K. Sinop.
\newblock Exphormer: Sparse transformers for graphs, 2023.

\bibitem[Sobolevsky(2021)]{sobolevsky2021hierarchical}
S.~Sobolevsky.
\newblock Hierarchical graph neural networks, 2021.

\bibitem[Southern et~al.(2024)Southern, Giovanni, Bronstein, and Lutzeyer]{southern2024vnbron}
J.~Southern, F.~D. Giovanni, M.~Bronstein, and J.~F. Lutzeyer.
\newblock Understanding virtual nodes: Oversmoothing, oversquashing, and node heterogeneity, 2024.

\bibitem[Topping et~al.(2022)Topping, Giovanni, Chamberlain, Dong, and Bronstein]{topping2022understanding}
J.~Topping, F.~D. Giovanni, B.~P. Chamberlain, X.~Dong, and M.~M. Bronstein.
\newblock Understanding over-squashing and bottlenecks on graphs via curvature, 2022.

\bibitem[Tönshoff et~al.(2023)Tönshoff, Ritzert, Rosenbluth, and Grohe]{toenshoff2023wheredid}
J.~Tönshoff, M.~Ritzert, E.~Rosenbluth, and M.~Grohe.
\newblock Where did the gap go? reassessing the long-range graph benchmark, 2023.

\bibitem[Xu et~al.(2019)Xu, Hu, Leskovec, and Jegelka]{xu2019powerful}
K.~Xu, W.~Hu, J.~Leskovec, and S.~Jegelka.
\newblock How powerful are graph neural networks?, 2019.

\bibitem[Ying et~al.(2019)Ying, You, Morris, Ren, Hamilton, and Leskovec]{ying2019hierarchical}
R.~Ying, J.~You, C.~Morris, X.~Ren, W.~L. Hamilton, and J.~Leskovec.
\newblock Hierarchical graph representation learning with differentiable pooling, 2019.

\bibitem[Zhang et~al.(2022)Zhang, Liu, Hu, and Lee]{zhang2022ansgt}
Z.~Zhang, Q.~Liu, Q.~Hu, and C.-K. Lee.
\newblock Hierarchical graph transformer with adaptive node sampling, 2022.

\bibitem[Zhu et~al.(2023)Zhu, Wen, Song, Ma, and Wang]{zhu2023hsgt}
W.~Zhu, T.~Wen, G.~Song, X.~Ma, and L.~Wang.
\newblock Hierarchical transformer for scalable graph learning, 2023.

\end{thebibliography}

\newpage
\appendix

\section{Properties of Hierarchical Support Graphs}
\subsection{Proof of Theorem~\ref{thm:basic-props}}
\label{sec:proof-basic-props}
\begin{proof}
    We denote the original graph as $G$ with $n$ nodes and $m$ edges. We further call $G^H$ the HSG-augmented graph. Clearly, for a constant coarsening ratio $r\in [0, 1)$, the number of nodes is the limit of a geometric series, yielding at most $\frac{n}{1-r}$ nodes in $G^H$. For a constant coarsening ratio, it is easy to prove that for appropriate $n$ the graph contains at most $\frac{\log n}{-\log r}$ layers, yielding a graph diameter of $2\frac{\log n}{-\log r}$. Based on the number of layers, the number of edges in each layer is at most $m$, yielding an upper bound $\mathcal{O}(\frac{m\log n}{-\log r})$.
\end{proof}

\subsection{Proof of Theorem~\ref{thm:const-ratio}}
\label{app:useful-coar}

\begin{proof}
    We define $r(i)$ to be the node reduction factor and $c(i)$ to be the edge reduction factor from layer $i-1$ to layer $i$. We write $R(i)$ and $C(i)$ as short form for the cumulative contraction until layer $i$ The average node degree in one support layer $H^{(i)}$ can thus be expressed as
    \begin{equation}
        \frac{C(i)m + R(i-1)n + R(i)n}{R(i)n}=\frac{C(i)m}{R(i)n} + \frac{1}{r(i)} + 1
    \end{equation}
    For this expression to have order $\Th(\frac{m}{n})$ one gets
    \begin{align}
        r(i) &= \Upomega\left(\frac{n}{m}\right)\\
        C(i) &\leq \Th(R(i))
    \end{align}
    If $m=o(n)$, meaning for sufficiently large $n$ the graph is disconnected, the bound on $r$ becomes $r=\Upomega(1)$.
\end{proof}

\section{Random Coarsening of Erdős–Rényi Graphs}
\label{app:random-proof}
As shown in Appendix~\ref{app:useful-coar} the conditions for a useful coarsening are restrictive in the coarsening ratio $r$ and in the edge reduction factor $c$. In this section, we investigate to what extent these properties are satisfied for a random coarsening on an Erdős-Rényi Graph.

\subsection{Proof of Theorem~\ref{thm:gnp-pvr}}
\begin{proof}
    We employ the common notation that $\sim$ denotes equality up to a multiplicative constant. \\
    For clusters $K_1, \dots, K_{r\cdot n}$ we calculate the expected number of intercluster edges between super-nodes for $i\neq j$ as
    \begin{equation}
        P[e(K_i,K_j)] = 1 - (1-p)^{\frac{1}{r^2}}\;.
    \end{equation}
    Clearly, in the next layer one has created a new Erdős–Rényi Graph with $G(rn, 1 - (1-p)^{\frac{1}{r^2}})$.
    We investigate the ratio of expected horizontal node degrees between the new layer $\E[d^{(1)}]$ and the original layer $\E[d^{(0)}]$
    \begin{equation}
    \label{eq:elem-ratio}
    \frac{\E[d^{(1)}]}{\E[d^{(0)}]} 
    \sim \frac{r(1 - (1-p)^{\frac{1}{r^2}})}{p}
    \end{equation}

    Given $p=n^{-1-\alpha}$ for $\alpha \geq 0$, $\E[m]={n\choose 2} p\sim n^{1-\alpha}$. Thus one finds the necessary coarsening ratio
    \begin{equation}
        r=\Th\left(\frac{n}{n+m}\right)=\Th\left(\frac{n}{n^{1-\alpha}+n}\right)=\Th(1)\;.
    \end{equation}
    As a result Equation~\ref{eq:elem-ratio} becomes
    \begin{equation}
        \frac{r(1 - (1-p)^{\frac{1}{r^2}})}{p} \sim n^{1+\alpha}(1 - (1-n^{-1-\alpha})^{\frac{1}{r^2}}) \sim 1
    \end{equation}
    Thus, sufficiently sparse graphs can be coarsened with a constant coarsening ratio. \\

    For $p=n^{-1+\alpha}$ and $\alpha > 0$ one gets $r\sim n^{-\alpha}$ and finds
    \begin{equation}
    \label{eq:basic-rel}
        \frac{r(1 - (1-p)^{\frac{1}{r^2}})}{p} \sim n^{1-2\alpha}\left(1-(1-n^{-1+\alpha})^{\left(n^{2\alpha}\right)}\right)
    \end{equation}

    For $\alpha=\frac{1}{2}$ the outer term becomes one, and the inner term converges to a constant. We give the following lower bound for the other cases:
    \begin{equation}
        n^{1-2\alpha}\left(1-(1-n^{-1+\alpha})^{\left(n^{2\alpha}\right)}\right)\geq n^{1-2\alpha}\left(1-\exp\left(-n^{-1+3\alpha}\right)\right)
    \end{equation}
    We make a case distinction on $\alpha$ and first consider the case $\alpha\in[\frac{1}{3}, \frac{1}{2})$. Since $\alpha\geq\frac{1}{3}$ the exponential term goes to zero, meaning the lower bound diverges. \\
    We next consider $\alpha\in(0, \frac{1}{3})$. By applying L'Hôpital's rule one can show that the lower bound also diverges in this interval.\\
    Finally, for $\frac{1}{2} < \alpha \leq 1$ the outer $n^{1-2\alpha}$ in Equation~\ref{eq:basic-rel} converges to zero, while the inner probability term can be upper-bounded by 1. \\
    
    Thus, the average horizontal node degree in the next layer remains unchanged up to constant factors only for $p=\Th(n^{-\beta})$ and $\beta\in\{\frac{1}{2}\}\cup[1,\infty)$. For $\beta\in[0,\frac{1}{2})$ the horizontal node degree goes to zero.
\end{proof}

\subsection{Proof of Theorem~\ref{thm:gnp-whatgraphs}}
\begin{proof}
    It follows directly from Theorem~\ref{thm:gnp-pvr} that values of $\beta$ for which the horizontal node degree diverges to infinity are not candidates for a node degree preserving coarsening. By assumption, the average node degree with respect to vertical edges must lie within $\Th(\frac{n}{m+n})$. As a result, all values of $\beta$ for which the node degree ratio as defined in Equation~\ref{eq:elem-ratio} does not diverge to infinity are permissible. This yields the original statement.
\end{proof}

\section{Model Configurations}
\label{app:model-configs}
We base our implementation on the GraphGPS \cite{rampasek2023graphgps} codebase, and integrate the coarsening approach from \cite{zhu2023hsgt}, which itself is based on the ClusterGCN implementation \cite{Chiang2019clustergcn}. We reuse several useful code snippets from the codebases of \cite{toenshoff2023wheredid,rosenbluth2024distinguished}. The models trained on the LRGB benchmark have less than 500K parameters, while models trained on ogbg-molpcba are not constrained to a parameter limit. \\
In Tables~\ref{tab:gcn-config} and~\ref{tab:gatedgcn-config} the row \textit{HSG node feats} denotes whether the node features of the created super-nodes are imputed or set to dummy edges. The same holds for \textit{HSG edge feats}. The \textit{pooling} row only refers to graph-level tasks and denotes whether standard global pooling or top layer pooling is used as introduced in Section~\ref{sec:adapted-pooling}. \textit{PE/SE} denotes which positional or structural encodings are used. LapPE denotes Laplacian Positional Encodings, and RWSE random walk structural encodings.
All models are trained using a learning rate of 0.001.

\begin{table}\scriptsize
    \centering
    \caption{Hyperparameter configurations for GCN-HSG on all tested datasets}
    \begin{tabular}{cccccc}
        \toprule
        hyperparam. & PascalVOC-SP & COCO-SP & Peptides-func & Peptides-struct & ogbg-molpcba\\
        \midrule
        dropout & 0.2 & 0.1 & 0.1 & 0.1 & 0.4\\
        num. layers & 10 & 14 & 6 & 12 & 4\\
        embed dim. & 140 & 200 & 235 & 105 & 1024\\
        pooling & - & - & global & top layer & top layer\\
        head depth & 2 & 1 & 3 & 2 & 1\\
        coarsening & $(\bullet)$ & $(0.05,\bullet)$ & $(0.5,\bullet)$ & $(\bullet)$ & $(\bullet)$\\
        HSG node feats & dummy & mean & dummy & dummy & dummy\\
        HSG edge feats & dummy & mean & dummy & dummy & dummy\\
        PE/SE & none & none & none & LapPE & RWSE\\
        batch size & 50 & 256 & 200 & 200 & 512\\
        num. epochs & 200 & 200 & 250 & 250 & 75\\
        num. params & 223K & 475K & 492K & 164K & 4.5M\\
        \bottomrule
    \end{tabular}
    \label{tab:gcn-config}
\end{table}

\begin{table}\scriptsize
    \centering
    \caption{Hyperparameter configurations for GatedGCN-HSG on all tested datasets}
    \begin{tabular}{cccccc}
        \toprule
        hyperparam. & PascalVOC-SP & COCO-SP & Peptides-func & Peptides-struct & ogbg-molpcba\\
        \midrule
        dropout & 0.2 & 0.1 & 0.1 & 0.1 & 0.4\\
        num. layers & 10 & 12 & 10 & 9 & 6\\
        embed dim. & 95 & 85 & 80 & 100 & 1024\\
        pooling & - & - & global & top layer & top layer \\
        head depth & 2 & 1 & 2 & 2 & 1\\
        coarsening & $(0.05, \bullet)$ & $(0.05,\bullet)$ & $(\bullet)$ & $(\bullet)$ & $(\bullet)$\\
        HSG nodes & mean & mean & dummy & dummy & dummy\\
        HSG edges & dummy & mean & dummy & dummy & dummy\\
        PE/SE & none & none & LapPE & LapPE & RWSE\\
        batch size & 50 & 256 & 200 & 200 & 512\\
        num. epochs & 200 & 200 & 250 & 250 & 75\\
        num. params & 473K & 452K & 348K & 486K & 31.8M\\
        \bottomrule
    \end{tabular}
    \label{tab:gatedgcn-config}
\end{table}
\end{document}